\documentclass[runningheads]{llncs}

\usepackage[T1]{fontenc}
\usepackage{graphicx,booktabs,tabularx,xcolor,hyperref}
\usepackage{amsmath,subcaption,float,amsfonts,amssymb,bm,mathtools,booktabs,threeparttable,nccmath}
\usepackage[title]{appendix}
\hypersetup{colorlinks=true,linkcolor=[rgb]{0.1,0.3,0.9},urlcolor=[rgb]{0.2,0.2,0.2},citecolor=[rgb]{0.1,0.3,0.9}}
\usepackage{url}

\begin{document}

\title{Robust Deep Learning for Myocardial Scar Segmentation in Cardiac MRI with Noisy Labels}
\titlerunning{Myocardial Scar Segmentation}

\author{Aida Moafi\inst{1,2} \and Danial Moafi\inst{3} \and Evgeny M. Mirkes\inst{4} \and Gerry P. McCann\inst{1,2} \and Abbas S. Alatrany\inst{1,2} \and Jayanth Ranjit Arnold\inst{1,2\star} \and Mostafa Mehdipour Ghazi\inst{5}\thanks{Co-senior corresponding authors: jra14@leicester.ac.uk and ghazi@di.ku.dk}}

\authorrunning{A. Moafi et al.}

\institute{
Department of Cardiovascular Sciences, University of Leicester, Leicester, UK \and
NIHR Leicester Biomedical Research Centre, Glenfield Hospital, Leicester, UK \and
Department of Information Engineering and Mathematics, University of Siena, Italy \and
Department of Mathematics, University of Leicester, Leicester, UK \and
Department of Computer Science, University of Copenhagen, Copenhagen, DK
}

\maketitle

\begin{abstract} % 150--250 words

The accurate segmentation of myocardial scars from cardiac MRI is essential for clinical assessment and treatment planning. In this study, we propose a robust deep-learning pipeline for fully automated myocardial scar detection and segmentation by fine-tuning state-of-the-art models. The method explicitly addresses challenges of label noise from semi-automatic annotations, data heterogeneity, and class imbalance through the use of Kullback-Leibler loss and extensive data augmentation. We evaluate the model’s performance on both acute and chronic cases and demonstrate its ability to produce accurate and smooth segmentations despite noisy labels. In particular, our approach outperforms state-of-the-art models like nnU-Net and shows strong generalizability in an out-of-distribution test set, highlighting its robustness across various imaging conditions and clinical tasks. These results establish a reliable foundation for automated myocardial scar quantification and support the broader clinical adoption of deep learning in cardiac imaging. The codes are available at: \url{https://github.com/Danialmoa/YoloSAM}.


\end{abstract}

\section{Introduction}

Cardiovascular disease remains a global health concern, responsible for nearly 32\% of all deaths worldwide \cite{lindstrom2022global}. Among its manifestations, myocardial infarction (MI) is a major cause of morbidity and mortality, often leaving patients with permanent myocardial scarring. Scar extent is a critical prognostic factor, with even a 5\% increase in scar size raising mortality risk by 19\% \cite{stone2016relationship}. The scar burden also impacts revascularization outcomes and predicts the likelihood of ventricular arrhythmias and heart failure \cite{kancharla2016scar,disertori2016myocardial}. Moreover, scar transmurality is essential for assessing recovery potential, as regions with less than 50\% transmural scar are more likely to regain function \cite{choi2001transmural,enein2020association,holtackers2022late}. Given its clinical importance, accurate scar quantification is crucial for personalized treatment and risk stratification.

Late gadolinium enhancement cardiovascular magnetic resonance imaging (LGE-CMR) is the gold standard for noninvasive scar assessment, offering high-resolution, contrast-enhanced visualization of infarcted tissue. However, current segmentation methods, including manual contouring and semi-automatic thresholding, are time-consuming, labor-intensive, and prone to inter-observer variability \cite{mewton2011assessment,khan2015comparison,kramer2020standardized}. Manual annotation can take up to 25 minutes per scan \cite{mewton2011assessment}, limiting scalability in clinical practice. Semi-automatic methods like the signal threshold-to-reference mean (STRM) and full width at half-maximum (FWHM) techniques segment scar tissue based on intensity thresholds but are sensitive to variations in signal-to-noise ratio, region-of-interest selection, and infarct signal characteristics \cite{klem2016sources,heiberg2022infarct,flett2011evaluation}. These limitations often lead to inconsistent scar delineation.

Deep learning offers promising solutions for automated scar segmentation \cite{chen2020deep}. Approaches based on 2D and 3D U-Net architectures \cite{ronneberger2015u} have demonstrated feasibility, as seen in the EMIDEC Challenge \cite{lalande2022deep}, but challenges related to ground-truth variability and imaging protocol differences hinder clinical adoption \cite{hoh2024impact}. A systematic review reported Dice similarity scores up to 0.63 for these methods \cite{jathanna2021diagnostic}, while manual annotation itself shows limited consistency, with intra- and inter-observer Dice indices of 0.76 and 0.69, respectively \cite{lalande2020emidec}. However, variability in LGE sequences and signal intensity further complicates segmentation. Magnitude and phase-sensitive inversion recovery (PSIR) sequences differ in spatial resolution and contrast properties, with PSIR offering better contrast but introducing inconsistencies in absolute intensity values \cite{stirrat2015influence,schulz2020standardized}. Differences in scanner vendors, acquisition parameters, and post-processing techniques add to the heterogeneity, complicating model generalization. Moreover, the small size and sporadic appearance of scars relative to the overall image can lead to model overfitting on background pixels. The quality of ground-truth labels significantly affects performance, as models often train on manual or semi-automatic annotations with inherent biases. Threshold-based methods like STRM and FWHM, while improving reproducibility, may not fully capture scar extent in cases of diffuse fibrosis, microvascular obstruction (MVO), or signal heterogeneity \cite{hennemuth2008comprehensive}. These challenges underscore the need for robust, generalizable deep learning models for automated myocardial scar segmentation.

We develop a deep-learning pipeline for fully automated segmentation of myocardial scars by fine-tuning state-of-the-art models \cite{tian2025yolov12,kirillov2023segment}. Our contributions are six-fold: (1) We propose an end-to-end pipeline achieving high-accuracy segmentation, producing smoother results despite being trained on noisy labels. (2) We introduce the Kullback-Leibler (KL) loss to mitigate the impact of imperfect manual or semi-automatic annotations. (3) We highlight the limitations of overlap-based metrics for evaluating small lesions and propose perimeter similarity as a more reliable size-based metric. (4) We use data augmentation and detection models to guide the segmentation process, addressing class imbalance and label noise. (5) We validate our approach on a heterogeneous, multi-cohort dataset, demonstrating its robustness across diverse imaging conditions. (6) We evaluate segmentation accuracy on an out-of-distribution test set, including MI scars, demonstrating generalizability to unseen data with a different task.
% To improve generalizability, the model leverages data augmentation to address data heterogeneity and class imbalance, key factors for improving accuracy, reproducibility, and clinical adoption of automated scar quantification.

\section{Methods}

\subsection{Data Acquisition and Annotation}

This study leverages a multi-cohort dataset comprising four independent studies on chronic reperfused ischemic scars, summarized in Table \ref{table-data}. The datasets were collected from multiple centers, encompassing diverse MRI acquisition protocols with variations in imaging sequences (e.g., magnitude and phase), scanner manufacturers (e.g., Siemens and Philips), magnetic field strengths (e.g., 1.5T and 3T), and spatial resolutions (e.g., from 1$\times$1$\times$6 mm$^3$ to 2$\times$2$\times$10 mm$^3$). These variations introduce heterogeneity in image quality, intensity, and contrast, posing challenges for consistent segmentation performance. The PSIR images were used when standard LGE images showed inconsistencies, as they are less dependent on inversion time selection and operator adjustments during acquisition. This approach enhanced the reliability of myocardial structure identification.

\begin{table}[b!]
\centering
\small
\caption{Summary of the multi-cohort dataset \cite{anonym} used in the study.}
\vspace{0.1cm}
\label{table-data}
\renewcommand{\arraystretch}{1.0}
\centering
\begin{tabular}{lcccc}
\toprule
 & \# Images & \# Subjects & Age & Sex \\
 & Train / Valid / Test \; & Train / Valid / Test \; & Mean $\pm$ SD \; & Female \\
\midrule
Cohort 1 \; & 450 / 74 / 91 & 49 / 9 / 11 & 59.44$\pm$9.85 & 25\% \\ % Dream
Cohort 2 \; & 493 / 82 / 88 & 59 / 12 / 13 & 45.41$\pm$21.20 & 65\% \\ % SCAD
Cohort 3 \; & 304 / 55 / 63 & 38 / 8 / 9 &  &  \\ % AMI
% Cohort 4 \; & 1063 / 168 & 120 / 20 &  &  \\ % Cvlprit
Cohort 4 \; & 471 / 54 / 41 & 49 / 8 / 6 &  &  \\ % Glenfield
Cohort 5 \; & 81 / 32 / 18 & 12 / 4 / 2 &  &  \\ % Harefield
Cohort 6 \; & 93 / 22 / 34 & 10 / 3 / 4 &  &  \\ % Kett
Cohort 7 \; & 129 / 27 / 26 & 15 / 3 / 3 &  &  \\ % Leeds
Cohort 8 \; & 117 / 37 / 49 & 12 / 4 / 5 &  &  \\ % Southampton
\bottomrule
\end{tabular}
\end{table} 

A total of 348 patients were selected, with LGE scar segmentation performed manually using commercially available Circle CVI software. To ensure high-quality annotations, all segmentations were reviewed by two expert cardiac MRI specialists. For reproducibility, scar delineation was performed using the widely used FWHM method. As a threshold-based technique, the FWHM method has inherent limitations, including sensitivity to intensity variations and partial volume effects, affecting segmentation accuracy. The labeling process involved three key steps: (1) Myocardial contouring, where the epicardial and endocardial borders were manually traced to define the myocardial region of interest. (2) Scar core identification, in which an ROI was placed within the region of highest signal intensity in the myocardium, corresponding to the scar core. (3) Refinement and exclusion, where a final revision step was conducted to exclude potential imaging artifacts and ensure consistency across cases.

\subsection{Preprocessing and Augmentation}

The annotated slices were preprocessed, accounting for DICOM pixel spacing, and scaled to 256$\times$256 pixels. We implemented different augmentation techniques to enhance the model's robustness to anatomical variability and imaging artifacts. In addition to standard augmentations with spatial transformations like flipping, rotation, scaling, and deformation, we integrated medical image-specific techniques to simulate real-world variations in MRIs. These augmentations included gamma correction, adaptive histogram equalization, brightness adjustment, and Rician noise addition to mimic MRI-specific noise characteristics. Bounding box jittering was also introduced to simulate variability in manual annotations and enhance detection robustness.

To ensure a robust and unbiased evaluation, we partitioned the datasets at the patient level, separating data from each cohort into training (70\%), validation (15\%), and test (15\%) subsets. This prevented data leakage and ensured the subsets were representative of the different cohorts. Given the inherent class imbalance in scar segmentation, we stratified the subsets based on the total number of scar pixels per patient. This approach maintained the full distribution of scar burden across the subsets, preventing the underrepresentation of critical cases. Each LGE-MRI scan was then processed at the slice level, with individual slices serving as network input. Since many slices in the LGE series did not contain the myocardium, this approach helped mitigate background imbalance and ensured that only relevant regions were considered during model development.

\subsection{Detection and Segmentation with Noisy Labels}

\begin{figure*}[!t]
\centering
\vspace{0.3cm}
\includegraphics[width=0.89\textwidth]{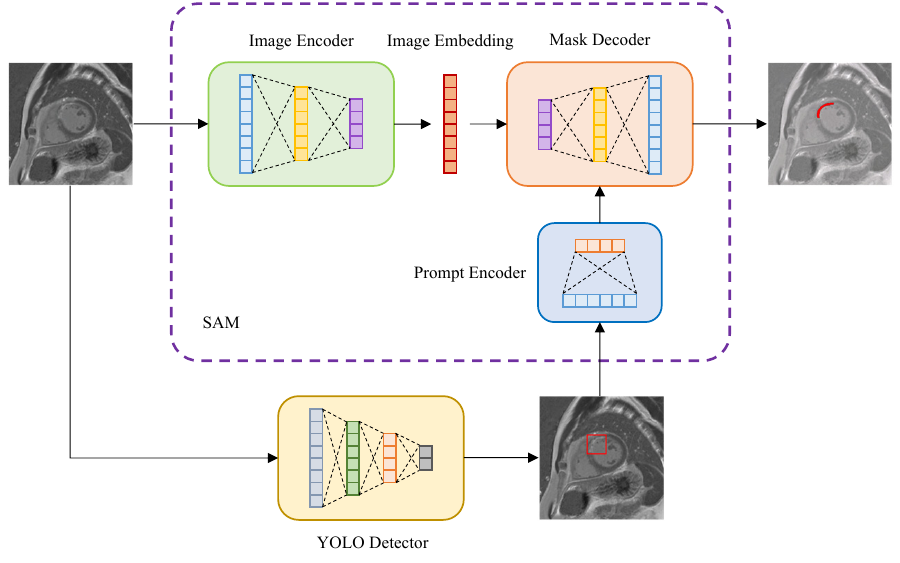}
\caption{Overview of the proposed fully automated pipeline for myocardial scar detection and segmentation. The YOLO model is fine-tuned to generate automatic bounding box prompts, guiding the fine-tuned SAM for precise segmentation of the localized scar. KL loss and extensive data augmentation strategies are employed to mitigate data variations and handle noisy label training.}
\label{figure-model}
\end{figure*}

Fig. \ref{figure-model} illustrates the fully automated pipeline developed in this study for detecting and segmenting myocardial scars with noisy ground truth labels. The pipeline employs the You Only Look Once (YOLO) architecture \cite{tian2025yolov12} with default configurations for scar bounding box detection, serving as prompts for the Segment Anything Model (SAM) \cite{kirillov2023segment}. This approach enables automated scar localization and provides a streamlined segmentation framework, eliminating the need for manual intervention. The detection step is crucial to focus on the tiny and fragmented scar regions. This method effectively integrates anatomical knowledge into deep learning, improving segmentation performance in challenging imaging scenarios and clinical applications where traditional methods struggle.

Despite reproducibility, the FWHM semi-automatic labels are sharp and often misaligned with true anatomical boundaries due to thresholding artifacts and partial volume effects. Hence, we consider them noisy in the sense of structured boundary uncertainty, rather than incorrect. To address the issue with noisy labels, we propose incorporating the KL divergence as an additional loss alongside the Dice loss and cross-entropy. We apply Gaussian smoothing with $\sigma$ = 2 to the target mask to obtain soft labels, capturing uncertainty in transitional boundary regions. The model predictions are mapped to [0, 1] using a sigmoid activation, followed by a log-softmax to obtain the predicted distribution. The prediction and the soft target scores are then normalized as probabilities, and KL divergence is applied between the probability distributions of predicted and soft target, penalizing overconfident predictions near ambiguous regions. The three losses are combined in a weighted manner, with weighting factors of 0.2, 0.2, and 0.6, respectively, determined by a grid search.

\subsection{Evaluation Metrics}

We evaluate segmentation performance using Hausdorff distance (HD), Dice similarity coefficient (DSC), area similarity (AS), and perimeter similarity (PS). Since DSC is an overlapping metric sensitive to small errors and may not fully capture performance on tiny, noisy (semi-automatic) scar labels, AS and PS provide complementary shape-based assessments. AS measures the pixel count similarity between ground truth ($T$) and prediction ($Y$) as $1 - ||Y| - |T|| / (|Y| + |T|)$, while PS applies the same formula to boundary pixel counts.

\section{Results}

\paragraph{\textbf{Baseline Comparison.}}
We first evaluated the nnU-Net \cite{isensee2021nnu} for segmentation, with results summarized in Table \ref{table-sota}. Next, we designed an experiment using ground-truth bounding boxes as prompts and fine-tuned SAM and modified DeepLabV3+ \cite{chen2018encoder} with a ResNet-50 backbone and additional input layers for handling bounding boxes. This approach significantly improved performance over the baseline nnU-Net, as shown in the table. These findings motivated the automation of SAM with YOLO-based bounding-box detection.

\begin{table*}[!t]
\centering
\small
\caption{Test segmentation accuracy (mean$\pm$SD) of the different models across all cohorts. Note that Box-SAM and Box-DeepLabV3+ refer to the models fine-tuned using ground-truth bounding boxes as input.}
\vspace{0.1cm}
\label{table-sota}
\renewcommand{\arraystretch}{1.25}
\centering
\begin{tabular}{lcccc}
\toprule
& DSC & HD & AS & PS \\
\bottomrule
nnU-Net \; & 0.579$\pm$0.338 \; & 11.786$\pm$16.004 \; & 0.702$\pm$0.345 \; & 0.748$\pm$0.341 \\
\midrule
Box-SAM \; & 0.653$\pm$0.135 \; & 7.807$\pm$7.922 \; & 0.793$\pm$0.133 \; & 0.918$\pm$0.080 \\
Box-DeepLabV3+ \; & 0.621$\pm$0.169 \; & 8.029$\pm$8.818 \; & 0.838$\pm$0.138 \; & 0.905$\pm$0.095 \\
\bottomrule
\end{tabular}
\end{table*}

\paragraph{\textbf{YOLO-SAM.}}
Table \ref{table-ours} presents the results of the proposed method based on the fine-tuned YOLO-SAM model. Although this fully automated pipeline introduces a slight performance drop, it eliminates the need for human intervention, unlike the semi-automatic approach that requires manual bounding box input.

\begin{table*}[!t]
\centering
\small
\caption{Test segmentation accuracy (mean$\pm$SD) of the proposed method (YOLO-SAM) across different cohorts.}
\vspace{0.1cm}
\label{table-ours}
\renewcommand{\arraystretch}{1.25}
\centering
\begin{tabular}{lcccc}
\toprule
& DSC & HD & AS & PS \\
\bottomrule
Cohort 1 \; & 0.579$\pm$0.338 \; & 11.786$\pm$16.004 \; & 0.702$\pm$0.345 \; & 0.748$\pm$0.341 \\
Cohort 2 \; & 0.636$\pm$0.258 \; & 9.362$\pm$8.469 \; & 0.814$\pm$0.243 \; & 0.865$\pm$0.232 \\
Cohort 3 \; & 0.525$\pm$0.280 \; & 15.993$\pm$18.012 \; & 0.753$\pm$0.301 \; & 0.782$\pm$0.299 \\
Cohort 4 \; & 0.579$\pm$0.406 \; & 8.931$\pm$13.915 \; & 0.628$\pm$0.412 \; & 0.670$\pm$0.418 \\
Cohort 5 \; & 0.596$\pm$0.308 \; & 15.741$\pm$15.535 \; & 0.779$\pm$0.317 \; & 0.805$\pm$0.271 \\
Cohort 6 \; & 0.516$\pm$0.373 \; & 15.388$\pm$16.646 \; & 0.636$\pm$0.373 \; & 0.680$\pm$0.373 \\
Cohort 7 \; & 0.666$\pm$0.377 \; & 13.217$\pm$24.006 \; & 0.767$\pm$0.368 \; & 0.774$\pm$0.357 \\
Cohort 8 \; & 0.628$\pm$0.361 \; & 7.599$\pm$11.248 \; & 0.734$\pm$0.347 \; & 0.778$\pm$0.339 \\
\midrule
Total \; & 0.601$\pm$0.330 \; & 10.728$\pm$14.217 \; & 0.753$\pm$0.318 \; & 0.797$\pm$0.311 \\
\bottomrule
\end{tabular}
\end{table*}

\begin{figure*}[!t]
\centering
\vspace{0.3cm}
\includegraphics[width=0.71\textwidth]{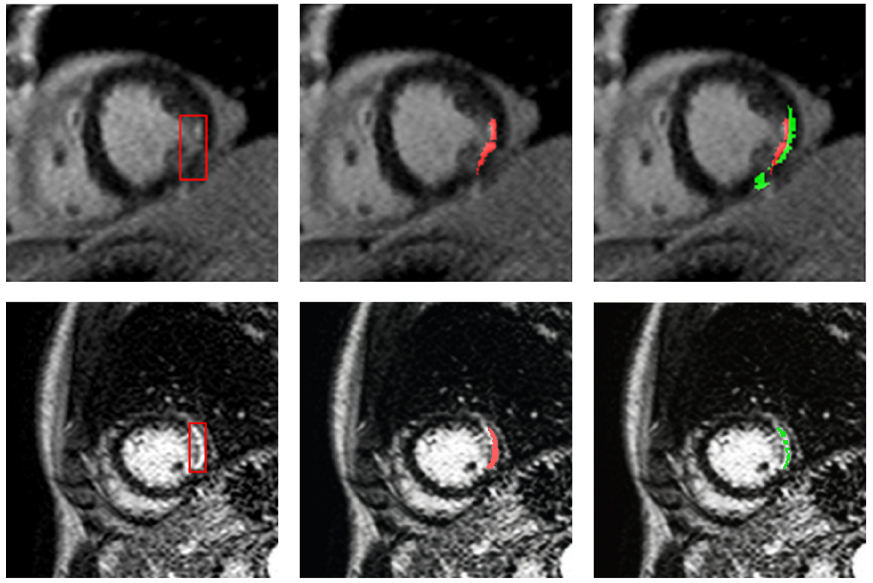}
\caption{Results of the proposed deep learning pipeline for myocardial scar segmentation. Each row shows a sample with three columns: (1) YOLO-based bounding box detection (red), (2) SAM-based automatic segmentation (red), and (3) semi-automatic FWHM-based ground truth (green). A slight spatial shift in the first predictions leads to a low Dice score of 0.15 despite a high perimeter similarity of 0.85, reflecting the model’s accurate scar extent but imprecise ground-truth localization. The second prediction is smooth, achieving a perimeter similarity of 0.98, though the fragmented ground truth results in a lower Dice score of 0.52.}
\label{figure-samples}
\end{figure*}

Fig. \ref{figure-samples} shows two segmented sample images using the proposed deep learning model compared to FWHM-based ground-truth labels. As seen, the automatic segmentation aligns with the ground-truth labels. However, a slight spatial shift in the predicted labels resulted in a very low Dice score of 0.15 due to the small overlap in the tiny scar regions in the first sample. This suggests that while the model captures the general extent of the scar, with a high perimeter similarity of 0.85, it struggles with precise localization. However, a closer inspection of the predicted labels reveals that they are often placed in more anatomically accurate locations. This observation highlights not only the limitations of the noisy, semi-automatically generated ground-truth labels but also the shortcomings of overlap-based metrics like the Dice score in evaluating segmentation accuracy. In contrast, size-based measures provide a more robust evaluation by assessing similarity independent of exact spatial alignment. Furthermore, the automatic predictions produce smoother and more coherent label regions than semi-automatic annotations, which often appear fragmented into small, scattered components. Despite the Dice score of 0.52 for the second sample, influenced by the discontinuity and small size of the ground-truth labels, the perimeter similarity remains high at 0.98, better reflecting the quality of the predicted segmentation.

We also emphasize the efficiency and clinical impact of predicted segmentations in scar detection and classification. To this end, we calculated the scar mass in the semi-automatic ground truth and the predicted segmentation in the test set. The estimated mass of the semi-automatic (ground truth) scars was 12.510$\pm$8.784 g, compared to 11.337$\pm$8.201 g for the fully-automatic (predicted) scars, indicating a very close estimation. The Wilcoxon rank sum test did not reveal a statistically significant difference between the estimated scar masses from the semi-automatic and predicted labels (\textit{p} = 0.42).
% Fig. \ref{figure-mass} presents violin plots comparing the estimated scar mass on predicted and semi-automatic labels using the FWHM method. While the average values and overall distributions across the test scans appear similar, there is a slight underestimation of the scar mass in the predicted segmentations. This may reflect a bias introduced by the semi-automatic and proposed segmentation algorithms.

\iffalse
\begin{figure*}[!t]
\centering
\vspace{0.3cm}
\includegraphics[width=0.7\textwidth]{fig_mass_pred.pdf}
\caption{Violin plots comparing the estimated scar mass between actual (semi-automatic) and predicted (automatic) segmentations across the test set.}
\label{figure-mass}
\end{figure*}
\fi

\paragraph{\textbf{Acute MI Generalization.}}

To assess generalization to different clinical domains, we evaluated the trained model in a multicenter acute MI cohort dataset with 50 patients and 406 images. The goal was to compare scar pattern differences between acute and chronic cases and to assess whether the predicted segmentations capture the expected clinical distinctions. Since the spatial characteristics of scars, including size, shape, and continuity, differ between acute and chronic cases, we extracted several morphological characteristics from the predicted scar regions. These features included scar size (number of pixels in the scar area), the number of separate lesions (number of connected components), compactness (solidity) defined as the ratio of the lesion area to its convex hull area, and elongation (circularity) calculated as $4\pi \,\cdot$ area / perimeter$^2$.

\begin{figure*}[!t]
\centering
\vspace{0.3cm}
\includegraphics[width=0.64\textwidth]{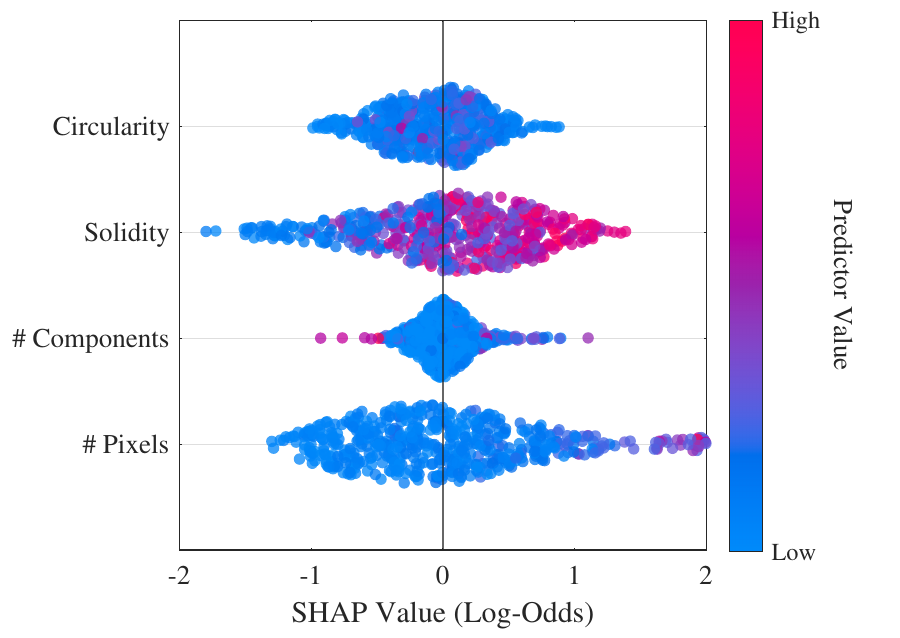}
\caption{Shapley summary plot for detecting acute MI cases based on morphological features extracted from the predicted scar segmentations.}
\label{figure-shaply}
\end{figure*}

To distinguish between acute and chronic scars, we trained an XGBoost classifier using these features extracted from the predicted segmentations. We then applied the Shapley method to explain the importance of each feature in the classification process. Fig. \ref{figure-shaply} shows the Shapley summary plot for detecting acute cases based on predicted scars, highlighting the greater burden and solidity of the infarct observed in acute cases. A statistical analysis of the extracted features further confirmed their significance in differentiating acute from chronic scars, with \textit{p}-values of 
4.8$\times$10$^{-7}$, 1.4$\times$10$^{-7}$, 1.7$\times$10$^{-4}$, and 
0.03 for scar size, solidity, circularity, and the number of connected components, respectively.

Using shape-based analysis of segmented scars offers the potential to develop diagnostic classifiers and predict patient risk and treatment outcomes. Such insights could enable cardiologists to rapidly distinguish between infarct types, which helps with both diagnosis and therapeutic decision-making. In particular, acute infarcts often exhibit regions of edema that may recover over time, whereas chronic infarcts typically represent irreversible damage. Identifying these distinct patterns could guide more effective treatment strategies. This experiment also underscored the advantage of training the model exclusively in chronic cases with more homogeneous and well-defined scar patterns, thus minimizing bias from the fragmented and less distinct appearance often seen in acute scars.

\section{Conclusion}

We developed a robust deep-learning pipeline for fully automated myocardial scar segmentation by fine-tuning state-of-the-art SAM and YOLO models. Our approach tackled key challenges in medical image analysis, including label noise from semi-automatic annotations, data heterogeneity, and class imbalance, using KL loss and extensive data augmentation strategies. Through a comprehensive evaluation of both acute and chronic cases, the model demonstrated its ability to produce smooth and accurate segmentations despite the presence of noisy labels.

The integration of a detection model to guide segmentation significantly improved accuracy by focusing on the scar region. Furthermore, the combination of diverse training data and systematic data augmentation enabled the model to generalize well to unseen test data with varying anatomical and pathological characteristics, as well as different acquisition parameters, including higher-resolution scans (e.g., 0.6$\times$0.6$\times$8 mm$^3$). Finally, the classification experiment using segmented acute MI scars provided a meaningful validation of the model’s ability to distinguish between acute and chronic cases, even when trained solely on chronic data. These findings underscore the potential of our approach for broader clinical application in automated cardiac imaging analysis.

\section*{Acknowledgments}
A.M. is funded by a University of Leicester Future 100 Studentship. M.M.G. is supported by the Pioneer Centre for AI (DNRF grant P1). J.R.A. is supported by a NIHR Clinician Scientist Award (CS-2018–18-ST2-007). G.P.M. is supported by a NIHR Research Professorship Award (RP-2017-08-ST2-007).

\bibliographystyle{splncs}
\bibliography{references}

\end{document}